\documentclass[11pt,a4paper]{article}
\usepackage[hyperref]{acl2019}
\usepackage{times}
\usepackage{latexsym}

\usepackage{url}

\usepackage{amsmath}

\usepackage{booktabs}
\usepackage{tikz}
\usepackage{amssymb}
\usepackage{mathtools}
\usepackage{algpseudocode}
\usepackage{algorithm}
\usepackage{euscript}
\DeclareMathAlphabet{\mathpzc}{T1}{pzc}{m}{it}
\usepackage{footnote}
\makesavenoteenv{tabular}
\makesavenoteenv{table}

\aclfinalcopy 
\def\aclpaperid{449} 

\usepackage{cleveref}
\crefname{section}{\S}{\S\S}
\Crefname{section}{?\S}{\S\S}

\DeclareMathOperator{\softmax}{softmax}
\DeclareMathOperator{\embed}{embed}
\DeclareMathOperator{\dropout}{dropout}
\DeclareMathOperator{\longestdist}{ldist}
\DeclareMathOperator{\topoorder}{topologic-order}
\DeclareMathOperator{\FF}{FF}

\DeclareMathOperator{\mypred}{\EuScript{R}^{--}}
\DeclareMathOperator{\mysucc}{\EuScript{R}^+}

\DeclareMathOperator{\mydirsucc}{\EuScript{N}^+}
\DeclareMathOperator{\mypredg}{\EuScript{R}^{--}_G}
\DeclareMathOperator{\mysuccg}{\EuScript{R}^+_G}
\DeclareMathOperator{\mydirpredg}{\EuScript{N}^{--}_G}
\DeclareMathOperator{\mydirsuccg}{\EuScript{N}^+_G}

\def\checkmark{\tikz\fill[scale=0.4](0,.35) -- (.25,0) -- (1,.7) -- (.25,.15) -- cycle;}

\title{Self-Attentional Models for Lattice Inputs}

       \author{Matthias Sperber$^1$, Graham Neubig$^2$, Ngoc-Quan Pham$^1$, Alex Waibel$^{1,2}$ 
       		\\ $^1$Karlsruhe Institute of Technology, Germany
		\\ $^2$Carnegie Mellon University, USA
		\\ \texttt{\{first\}.\{last\}@kit.edu, gneubig@cs.cmu.edu}
		}

\date{}

\begin{document}
\maketitle
\begin{abstract}

Lattices are an efficient and effective method to encode ambiguity of upstream systems in natural language processing tasks, for example to compactly capture multiple speech recognition hypotheses, or to represent multiple linguistic analyses. Previous work has extended recurrent neural networks to model lattice inputs and achieved improvements in various tasks, but these models suffer from very slow computation speeds. This paper extends the recently proposed paradigm of self-attention to handle lattice inputs. Self-attention is a sequence modeling technique that relates inputs to one another by computing pairwise similarities and has gained popularity for both its strong results and its computational efficiency. To extend such models to handle lattices, we introduce probabilistic reachability masks that incorporate lattice structure into the model and support lattice scores if available. We also propose a method for adapting positional embeddings to lattice structures. We apply the proposed model to a speech translation task and find that it outperforms all examined baselines while being much faster to compute than previous neural lattice models during both training and inference.
\end{abstract}

\section{Introduction}

\begin{figure}[t]
\centering
\includegraphics[width=4.5cm]{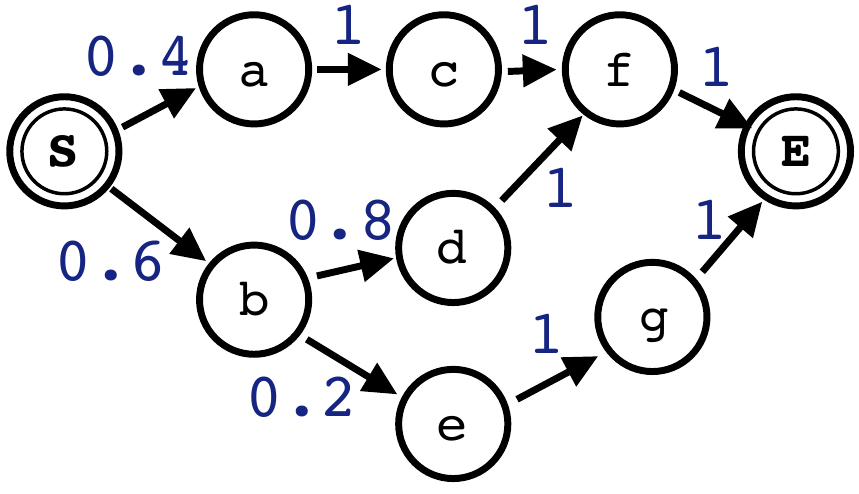}
\caption{Example of a node-labeled lattice. Nodes are labeled with word tokens and posterior scores.}
\label{fig:example-lattice}
\end{figure}

In many natural language processing tasks, graph-based representations have proven useful tools to enable models to deal with highly structured knowledge. Lattices are a common instance of graph-based representations that allows capturing a large number of alternative sequences in a compact form (Figure~\ref{fig:example-lattice}). Example applications include speech recognition lattices that represent alternative decoding choices \cite{Saleem2004,Zhang2005,Matusov2008}, word segmentation lattices that capture ambiguous decisions on word boundaries or morphological alternatives \cite{Dyer2008}, word class lattices \cite{Navigli2010}, and lattices for alternative video descriptions \cite{Rohrbach2014}.

Prior work has made it possible to handle these through the use of recurrent neural network (RNN) lattice representations \cite{Ladhak2016,Su2017,Sperber2017}, inspired by earlier works that extended RNNs to tree structures \cite{Socher2013,Tai2015,Zhu2015a}. Unfortunately, these models are computationally expensive, because the extension of the already slow RNNs to tree-structured inputs prevents convenient use of batched computation. An alternative model, graph convolutional networks (GCN) \cite{Duvenaud2015,Defferrard2016,kearnes2016molecular,kipf2016semi}, is much faster but considers only local context and therefore requires combination with slower RNN layers for typical natural language processing tasks \cite{Bastings2017,Cetoli2017,Vashishth2018}.

For linear sequence modeling, self-attention \cite{cheng2016a,Parikh2016,Lin2017,Vaswani2017} now provides an alternative to RNNs. Self-attention encodes sequences by relating sequence items to one another through computation of pairwise similarity, with addition of positional encoding to model positions of words in a linear sequence. Self-attention has gained popularity thanks to strong empirical results and computational efficiency afforded by parallelizable computations across sequence positions. 

In this paper, we extend the previously purely sequential self-attentional models to lattice inputs. Our primary goal is to obtain additional modeling flexibility while avoiding the increased cost of previous lattice-RNN-based methods. Our technical contributions are two-fold: First, we incorporate the global lattice structure into the model through reachability masks that mimic the pairwise conditioning structure of previous recurrent approaches. These masks can account for lattice scores if available. Second, we propose the use of lattice positional embeddings to model positioning and ordering of lattice nodes.

We evaluate our method on two standard speech translation benchmarks, replacing the encoder component of an attentional encoder-decoder model with our proposed lattice self-attentional encoder. Results show that the proposed model outperforms all tested baselines, including LSTM-based and self-attentional sequential encoders, a LatticeLSTM encoder, and a recently proposed self-attentional model that is able to handle graphs but only considers local context, similar to GCNs. The proposed model performs well without support from RNNs and offers computational advantages in both training and inference settings.

\section{Background}

\subsection{Masked Self-Attention}

We start by introducing self-attentional models for sequential inputs, which we will extend to lattice-structured inputs in \cref{sec:proposed_model}.

Attentional models in general can be described using the terminology of queries, keys, and values. The input is a sequence of $l$ values, along with a key corresponding to each value. For some given query, the model computes how closely each key matches the query. 
Here, we assume values, keys, and queries $\mathbf{v}_k,\mathbf{k}_k,\mathbf{q}{\in}\mathbb{R}^d$, for some dimensionality $d$ and sequence indices $k{\in}\{1\dots l\}$. Using the computed similarity scores $f(\mathbf{q},\mathbf{k}_k)$, attention computes a weighted average of the values to obtain a fixed-size representation of the whole sequence conditioned on this query. In the self-attentional case, the sequence items themselves are used as queries, yielding a new sequence of same length as output in which each of the original input elements has been enriched by the respectively relevant global context.

The following equations formalize this idea. We are given a sequence of input vectors $\mathbf{x}_k\in \mathbb{R}^d$. For every query index $i$, we compute an output vector $\mathbf{y}_i$ as:

\begin{align}
e_{ij}&=f\left(q\left(\mathbf{x}_i\right),k\left(\mathbf{x}_j\right)\right){+}m_{ij} & (\forall1{\le}j{\le}l)  \label{eq:background-sa-score}   \\
\boldsymbol{\alpha}_{i}&=\softmax\left( \mathbf{e}_{i} \right) \\ 
\mathbf{y}_i&=\sum_{j=1}^l \alpha_{ij}v\left(\mathbf{x}_j\right). \label{eq:weighted_sum}
\end{align}

Here, unnormalized pairwise similarities $e_{ij}$ are computed through the similarity function $f$, and then normalized as $\alpha_{ij}$ for computation of a weighted sum of value vectors. $q,k,v$ denote parametrized transformations (e.g.\ affine) of the inputs into queries, keys, and values.

Equation~\ref{eq:background-sa-score} also adds an attention masking term $m_{ij}\in\mathbb{R}$ that allows adjusting or disabling the influence of context at key position $j$ on the output representation at query position $i$. Masks have, for example, been used to restrict self-attention to ignore future decoder context \cite{Vaswani2017} by setting $m_{ij}={-}\infty$ for all $j{>}i$. We will use this concept in \cref{sec:reachability_masks} to model reachability structure.

\subsection{Lattices}
\label{sec:background-lattices}

We aim to design models for lattice inputs that store a large number of sequences in a compact data structure, as illustrated in  Figure~\ref{fig:example-lattice}. We define lattices as directed acyclic graphs (DAGs) with the additional property that there is exactly one start node (\texttt{S}) and one end node (\texttt{E}). We call the sequences contained in the lattice \textit{complete} paths, running from the start node to the end node.
Each node is labeled with a word token.\footnote{Edge-labeled lattices can be easily converted to node-labeled lattices using the line-graph algorithm \cite{Hemminger1978}.}

To make matters precise, let $G{=}(V,E)$ be a DAG with nodes $V$ and edges $E$. For $k{\in}V$, let $\mysuccg(k)$ denote all successors (reachable nodes) of node $k$, and let $\mydirsuccg(k)$ denote the neighborhood, defined as the set of all adjacent successor nodes. $\mypredg(k),\mydirpredg(k)$ are defined analogously for predecessors. $j\succ i$ indicates that node $j$ is a successor of node $i$.

For arbitrary nodes $i, j$, let $p_{G}\left(j\succ i\mid i\right)$ be the probability that a complete path in $G$ contains $j$ as a successor of $i$, given that $i$ is contained in the path. Note that $j\notin \mysuccg(i)$ implies $p_{G}\left(j\succ i\mid i\right){=}0$. 
The probability structure of the whole lattice can be represented through transition probabilities $p_{k,j}^{\text{trans}}{:=}p_{G}\left(k\succ j\mid j\right)$ for $j\in\mydirsuccg(k)$.
We drop the subscript $G$ when clear from context.

\section{Baseline Model}
\label{sec:baseline-model}

Our proposed model builds on established architectures from prior work, described in this section.

\subsection{Lattice-Biased Attentional Decoder}
\label{sec:enc-dec}

The common attentional encoder-decoder model \cite{Bahdanau2014} serves as our starting point. The encoder will be described in \cref{sec:proposed_model}. As cross-attention mechanism, we use the  lattice-biased variant \cite{Sperber2017}, which adjusts the attention scores $\alpha_{ij}^\text{cross}$ between encoder position $j$ and decoder position $i$ according to marginal lattice scores $p\left(j\succ \mathtt{S}\mid\mathtt{S}\right)$ (\cref{sec:prob_masks} describes how to compute these) as follows:\footnote{We have removed the trainable peakiness coefficient from the original formulation for simplicity and because gains of this additional parameter were unclear according to \newcite{Sperber2017}.}
\begin{align}
\alpha_{ij}^\text{cross}\propto\exp\left(\textit{score}(\bullet)+\log p\left(j\succ \mathtt{S}\mid\mathtt{S}\right)\right). \label{eq:cross-att}
\end{align}
Here, $\textit{score}(\bullet)$ is the unnormalized attention score.

In the decoder, we use long short-term memory (LSTM) networks, although it is straightforward to use alternative decoders in future work, such as the self-attentional decoder proposed by \newcite{Vaswani2017}. We further use input feeding \cite{Luong2015}, variational dropout in the decoder LSTM \cite{Gal2016}, and label smoothing \cite{Szegedy2016}.

\subsection{Multi-Head Transformer Layers}
\label{sec:transformer}
To design our self-attentional encoder, we use \newcite{Vaswani2017}'s \textit{Transformer} layers that combine self-attention with position-wise feed-forward connections, layer norm \cite{Ba2016a}, and residual connections \cite{He2016b} to form deeper models. Self-attention is modeled with multiple heads, computing independent self-attentional representations for several separately parametrized attention heads, before concatenating the results to a single representation. This increases model expressiveness and allows using different masks (Equation~\ref{eq:background-sa-score}) between different attention heads, a feature that we will exploit in \cref{sec:reachability_masks}. Transformer layers are computed as follows:

\begin{align}
\mathbf{Q}_{k}&=\mathbf{X}\mathbf{W}_{k}^\text{(q)},
\mathbf{K}_{k}{=}\mathbf{X}\mathbf{W}_{k}^\text{(k)},
\mathbf{V}_{k}{=}\mathbf{X}\mathbf{W}_{k}^\text{(v)}  \\
\mathbf{H}_{k}&=\softmax\left(\frac{1}{\sqrt{d}}\dropout\left(\mathbf{Q}_{i}\mathbf{K}_{k}^\top{+}\mathbf{M}\right)\right)\mathbf{V}_{k}  \label{eq:head}\\
\mathbf{H}&=\mathrm{concat}(\mathbf{H}_{1},\mathbf{H}_{2},\ldots,\mathbf{H}_n) \\
\mathbf{L}&=\mathrm{LN}\left[\dropout\left(\mathbf{H}+\mathbf{X}\right)\right] \\
\mathbf{Y}&=\mathrm{LN}\left[\dropout\left(\FF\left(\mathbf{L}\right)+\mathbf{L}\right)\right]
\end{align}
Here, $\mathbf{X}{\in}{\mathbb{R}^{l\times d}},\mathbf{Q}_k,\mathbf{K}_k,\mathbf{V}_k{\in}{\mathbb{R}^{l\times d/n}}$ denote inputs and their query-, key-, and value transformations for attention heads with index $k{\in}\{1,\dots,n\}$, sequence length $l$, and hidden dimension $d$. $\mathbf{M}{\in}{\mathbb{R}^{l\times l}}$ is an attention mask to be defined in \cref{sec:reachability_masks}. Similarity between keys and queries is measured via the dot product. The inputs are word embeddings in the first layer, or the output of the previous layer in the case of stacked layers. $\mathbf{Y}{\in}{\mathbb{R}^{l\times d}}$ denotes the final output of the Transformer layer. $\mathbf{W}_{k}^\text{(q)}, \mathbf{W}_{k}^\text{(k)}, \mathbf{W}_{k}^\text{(v)}\in\mathbb{R}^{d\times d/n}$ are parameter matrices. $\FF$ is a position-wise feed-forward network intended to introduce additional depth and nonlinearities, defined as $\FF(x){=}\max\left(\boldsymbol{0},\mathbf{x}\mathbf{W}_1+\mathbf{b}_1\right)\mathbf{W}_2+\mathbf{b}_2$. $\text{LN}$ denotes layer norm. Note that dropout regularization \cite{Srivastava2014} is added in three places.

Up to now, the model is completely agnostic of sequence positions. However, position information is crucial in natural language, so a mechanism to represent such information in the model is needed. A common approach is to add positional encodings to the word embeddings used as inputs to the first layer. We opt to use learned positional embeddings \cite{Gehring2017}, and obtain the following after applying dropout:
\begin{align}
\mathbf{x}_i'=\dropout\left(\mathbf{x}_i+\embed\left[i\right]\right). \label{eq:seq_pos_emb}
\end{align}
Here, a position embedding $\embed\left[i\right]$ of equal dimension with sequence item $\mathbf{x}_i$ at position $i$ is added to the input.

\section{Self-Attentional Lattice Encoders}
\label{sec:proposed_model}

A simple way to realize self-attentional modeling for lattice inputs would be to linearize the lattice in topological order and then apply the above model. However, such a strategy would ignore the lattice structure and relate queries to keys that cannot possibly appear together according to the lattice. We find empirically that this naive approach performs poorly (\cref{sec:ablation}). As a remedy, we introduce a masking scheme to incorporate lattice structure into the model (\cref{sec:reachability_masks}), before addressing positional encoding for lattices (\cref{sec:lattice_positions}).

\subsection{Lattice Reachability Masks}
\label{sec:reachability_masks}

We draw inspiration from prior works such as the TreeLSTM \cite{Tai2015} and related works. Consider how the recurrent conditioning of hidden representations in these models is informed by the graph structure of the inputs: Each node is conditioned on its direct predecessors in the graph, and via recurrent modeling on all its predecessor nodes up to the root or leaf nodes.

\subsubsection{Binary Masks}
\label{sec:bin-masks}

\begin{figure}[tb]
\centering
\includegraphics[width=8cm]{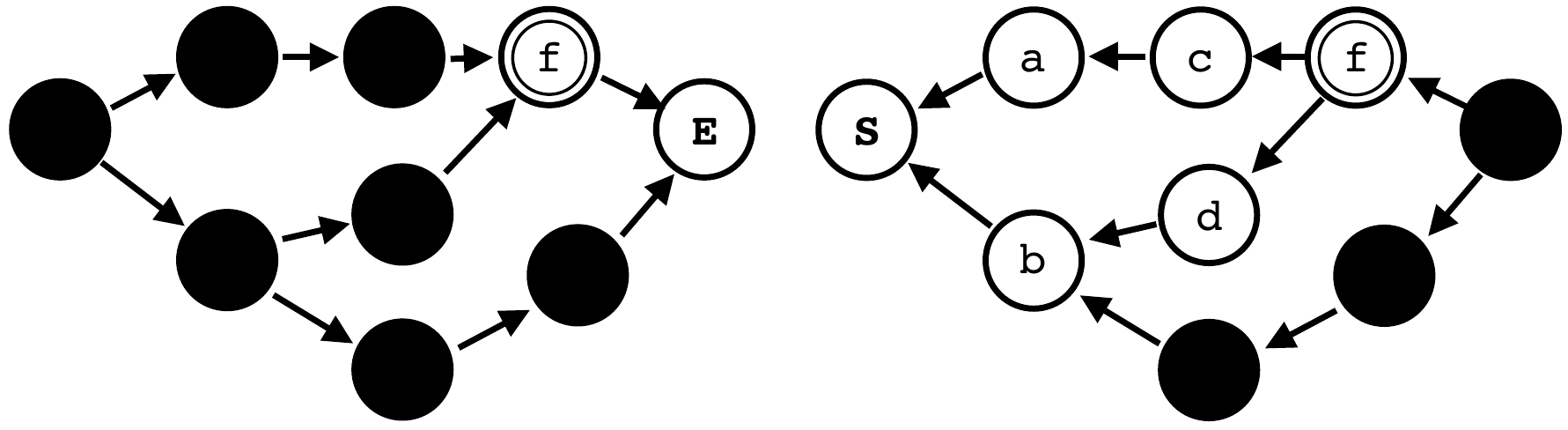}
\caption{Example for binary masks in forward- and backward directions. The currently selected query is node \texttt{f}, and the mask prevents all solid black nodes from being attended to.}
\label{fig:binary-mask}
\end{figure}

We propose a masking strategy that results in the same conditioning among tokens based on the lattice structure, preventing the self-attentional model from attending to lattice nodes that are not reachable from some given query node $i$. Figure~\ref{fig:binary-mask} illustrates the concept of such reachability masks. Formally, we obtain masks in forward and backward direction as follows:

\begin{align*}
  \overrightarrow{m}_{ij}^\text{bin}&=\left\{
            \begin{array}{ll}
            0 &\text{if } i{\in}\mypred\left(j\right)\lor i{=}j\\
            -\infty &\text{else}
            \end{array}
            \right.
\\  \overleftarrow{m}_{ij}^\text{bin}&=\left\{
            \begin{array}{ll}
            0 &\text{if } i{\in}\mysucc\left(j\right)\lor i{=}j\\
            -\infty &\text{else}
            \end{array}
            \right.
\end{align*}

The resulting conditioning structure is analogous to the conditioning in  lattice RNNs \cite{Ladhak2016} in the backward and forward directions, respectively. These masks can be obtained using standard graph traversal algorithms.

\subsubsection{Probabilistic Masks}
\label{sec:prob_masks}

\begin{figure}[tb]
\centering
\includegraphics[width=8cm]{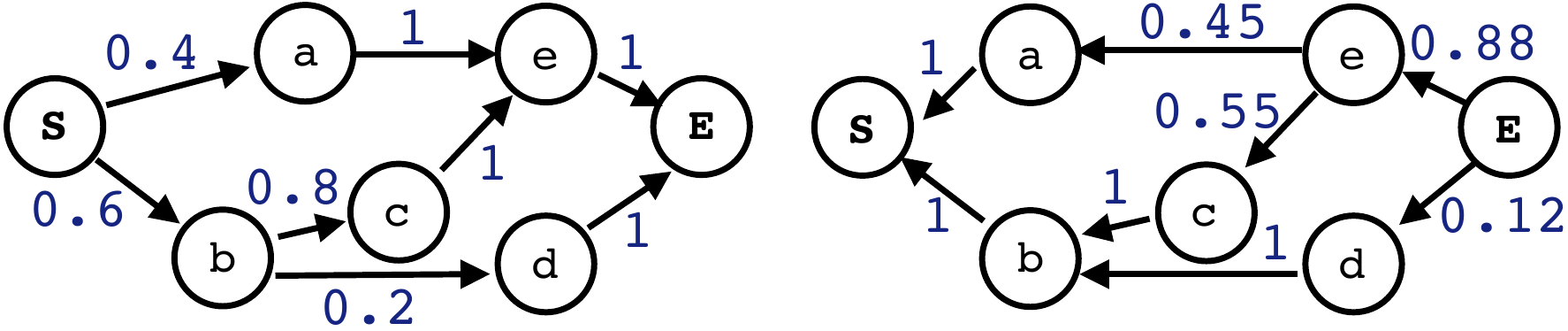}
{\small
\begin{tabular}{c|ccccccccc}
$\rightarrow$  & \texttt{S} & \texttt{a}   & \texttt{b}   & \texttt{c}   & \texttt{d}    & \texttt{e}    & \texttt{E} \\
\midrule
\texttt{S}         & 1            & 0.4            & 0.6          & 0.48          & 0.12           & 0.88           & 1       \\
\texttt{a}         & 0            & 1               & 0              & 1              & 0                & 1              & 1          \\
\texttt{b}          & 0           & 0               & 1              & 0.8           & 0.2             & 0.8           & 1          \\
\texttt{c}         & 0            & 0               & 0              & 1              & 0                & 1              & 1             \\
\texttt{d}         & 0            & 0               & 0              & 0              & 1                & 0              & 1             \\
\texttt{e}          & 0            & 0               & 0              & 0              & 0                & 1              & 1    \\
\texttt{E}         & 0            & 0               & 0              & 0              & 0                & 0              & 1
\end{tabular}
\begin{tabular}{c|ccccccccc}
$\leftarrow$  & \texttt{S} & \texttt{a}    & \texttt{b}    & \texttt{c}    & \texttt{d}   & \texttt{e}   & \texttt{E} \\
\midrule
\texttt{S}        & 1            & 0               &  0              & 0                & 0              & 0             & 0 \\
\texttt{a}        & 1            & 1               &  0              & 0                & 0              & 0             & 0 \\
\texttt{b}        & 1            & 0               &  1              & 0                & 0              & 0             & 0 \\
\texttt{c}        & 1            & 0               &  0              & 1                & 0              & 0             & 0 \\
\texttt{d}        & 1            & 0               &  1              & 0                & 1              & 0             & 0 \\
\texttt{e}         & 1            & 0.45         & 0.55          & 0.55           & 0              & 1             & 0 \\
\texttt{E}        & 1            & 0.4            &   0.6          & 0.48           & 0.12         &  0.88         & 1
\end{tabular}
}
\caption{Example for pairwise conditional reaching probabilities for a given lattice, which we logarithmize to obtain self-attention masks. Rows are queries, columns are keys.}
\label{fig:prob-mask}
\end{figure}

Binary masks capture the graph structure of the inputs, but do not account for potentially available lattice scores that associate lattice nodes with a probability of being correct. Prior work has found it critical to exploit lattice scores, especially for noisy inputs such as speech recognition lattices \cite{Sperber2017}. In fact, the previous binary masks place equal weight on all nodes, which will cause the influence of low-confidence regions (i.e., dense regions with many alternative nodes) on computed representations to be greater than the influence of high-confidence regions (sparse regions with few alternative nodes).

It is therefore desirable to make the self-attentional lattice model aware of these scores, so that it can place higher emphasis on confident context and lower emphasis on context with low confidence. The probabilistic masks below generalize binary masks according to this intuition:

\begin{align*}
  \overrightarrow{m}_{ij}^\text{prob}&=\left\{
            \begin{array}{ll}
            \log p_{G}\left(j\succ i\mid i\right) &\text{if } i{\neq}j\\
            0 & \text{if } i{=}j
            \end{array}
            \right.
\\  \overleftarrow{m}_{ij}^\text{prob}&=\left\{
            \begin{array}{ll}
            \log p_{G^\top}\left(j\succ i\mid i\right) &\text{if } i{\neq}j\\
            0 & \text{if } i{=}j
            \end{array}
            \right.
\end{align*}

Here, we set $\log(0){\coloneqq} {-}\infty$. Figure~\ref{fig:prob-mask} illustrates the resulting pairwise probability matrix for a given lattice and its reverse, prior to applying the logarithm. Note that the first row in the forward matrix and the last row in the backward matrix are the globally normalized scores of Equation~\ref{eq:cross-att}.

Per our convention regarding $\log(0)$, the $-\infty$ entries in the mask will occur at exactly the same places as with the binary reachability mask, because the traversal probability is 0 for unreachable nodes. For reachable nodes, the probabilistic mask causes the computed similarity for low-confident nodes (keys) to be decreased, thus increasing the impact of confident nodes on the computed hidden representations.
The proposed probabilistic masks are further justified by observing that the resulting model is invariant to path duplication (see 
Appendix~\ref{app:duplication-invariance}), unlike the model with binary masks.

The introduced probabilistic masks can be computed in $\mathcal{O}\left(|V|^3\right)$ from the given transition probabilities by using the dynamic programming approach described in Algorithm~\ref{algo:dp}. The backward-directed probabilistic mask can be obtained by applying the same algorithm on the reversed graph.

\begin{algorithm}[tb]
\caption{Computation of logarithmized probabilistic masks via dynamic programming. \\-- given: DAG $G=(V,E)$; transition probs $p_{k,j}^\text{trans}$}
\label{algo:dp}
\begin{algorithmic}[1]
\State $\forall i,j\in V: q_{i,j}\gets 0$
\For{$i\in V$} \Comment{loop over queries}
	\State $q_{i,i}\gets 1$
	\For{$k\in \topoorder\left(V\right)$}
		\For{next $\in\mydirsucc\left(k\right)$}
			\State $q_{i,\text{next}}\gets q_{i,\text{next}}+p_\text{k,next}^\text{trans}\cdot q_{i,k}$
		\EndFor
	\EndFor
\EndFor
\State $\forall i,j\in V: m_{ij}^\text{prob}\gets \log q_{i,j}$
\end{algorithmic}
\end{algorithm}

\subsubsection{Directional and Non-Directional Masks}

The above masks are designed to be plugged into each Transformer layer via the masking term $\mathbf{M}$ in Equation~\ref{eq:head}. However, note that we have defined two different masks, $\overrightarrow{m}_{ij}$ and $\overleftarrow{m}_{ij}$. To employ both we can follow two strategies: (1) Merge both into a single, {\it non-directional} mask by using $\overleftrightarrow{m}_{ij}=\max\left\{{\overrightarrow{m}_{ij},\overleftarrow{m}_{ij}}\right\}$. (2) Use half of the attention heads in each multi-head Transformer layer (\cref{sec:transformer}) with forward masks, the other half with backward masks, for a {\it directional} strategy.

Note that when the input is a sequence (i.e., a lattice with only one complete path), the non-directional strategy reduces to unmasked sequential self-attention. The second strategy, in contrast, reduces to the directional masks proposed by \newcite{Shen2018} for sequence modeling.

\subsection{Lattice Positional Encoding}
\label{sec:lattice_positions}

\begin{figure}[tb]
\centering
\includegraphics[width=5cm]{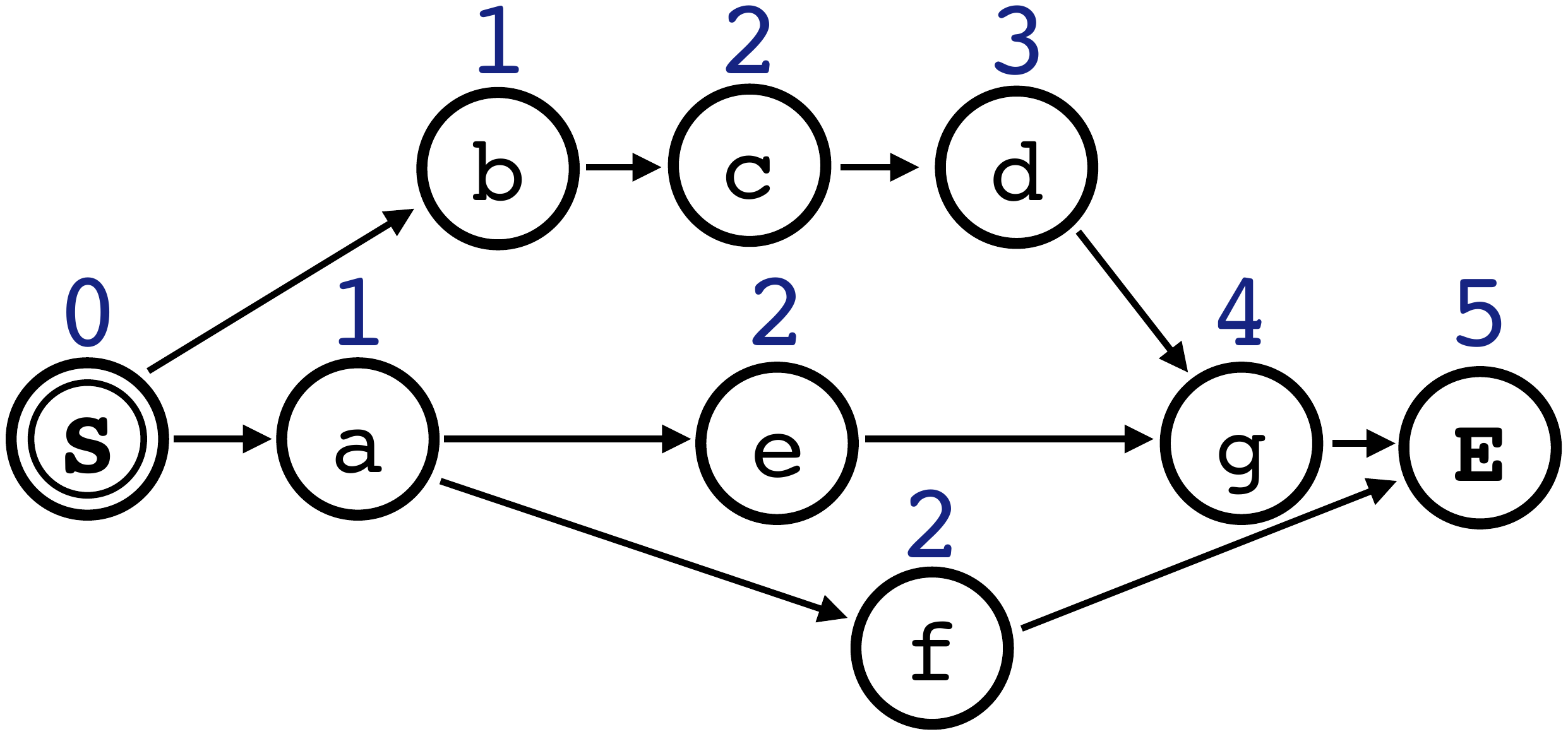}
\caption{Lattice positions, computed as longest-path distance from the start node \texttt{\textbf{S}}.}
\label{fig:lattice-pos}
\end{figure}

Encoding positional information in the inputs is a crucial component in self-attentional architectures as explained in \cref{sec:transformer}. To devise a strategy to encode positions of lattice nodes in a suitable fashion, we state a number of desiderata: (1) Positions should be integers, so that positional embeddings (\cref{sec:transformer}) can be used. (2) Every possible lattice path should be assigned strictly monotonically increasing positions, so that relative ordering can be inferred from positions. (3) For a compact representation, unnecessary jumps should be avoided. In particular, for at least one complete path the positions should increase by exactly 1 across all adjacent succeeding lattice nodes.

A naive strategy would be to use a topological order of the nodes to encode positions, but this clearly violates the compactness desideratum. \newcite{Dyer2008} used shortest-path distances between lattice nodes to account for distortion, but this violates monotonicity. Instead, we propose using the longest-path distance ($\longestdist$) from the start node, replacing Equation~\ref{eq:seq_pos_emb} with: $$\mathbf{x}_i'=\dropout\left(\mathbf{x}_i+\embed\left[\longestdist\left(\mathtt{S}\rightarrow i\right)\right]\right).$$ This strategy fulfills all three desiderata, as illustrated in Figure~\ref{fig:lattice-pos}. Longest-path distances from the start node to all other nodes can be computed in $\mathcal{O}\left(|V|^2\right)$ using e.g.\ Dijkstra's shortest-path algorithm with edge weights set to $-1$.

\subsection{Computational Complexity}
The computational complexity in the self-attentional encoder is dominated by generating the masks ($\mathcal{O}\left(|V|^3\right)$), or by the computation of pairwise similarities ($\mathcal{O}\left(|V|^2\right)$) if we assume that masks are precomputed prior to training. Our main baseline model, the LatticeLSTM, can be computed in $\mathcal{O}\left(|E|\right)$, where $|E|\le |V|^2$. Nevertheless, constant factors and the effect of batched operations lead to considerably faster computations for the self-attentional approach in practice (\cref{sec:experiments-speed}).

\section{Experiments}
\label{sec:experiments}


We examine the effectiveness of our method on a speech translation task, in which we directly translate decoding lattices from a speech recognizer into a foreign language.

\subsection{Settings}
\label{sec:settings}
We conduct experiments on the Fisher--Callhome Spanish--English Speech Translation corpus \cite{Post2013}. This corpus contains translated telephone conversations, along with speech recognition transcripts and lattices. The Fisher portion (138k training sentences) contains conversations between strangers, and the smaller Callhome portion (15k sentences) contains conversations between family members. Both and especially the latter are acoustically challenging, indicated by speech recognition word error rates of 36.4\% and 65.3\% on respective test sets for the transcripts contained in the corpus. The included lattices have oracle word error rates of 16.1\% and 37.9\%.

We use XNMT \cite{neubig18xnmt} which is based on DyNet \cite{Neubig2017}, with the provided self-attention example as a starting point.\footnote{Our code is available: \url{http://msperber.com/research/acl-lattice-selfatt/}} Hidden dimensions are set to 512 unless otherwise noted. We use a single-layer LSTM-based decoder with dropout rate 0.5. All self-attentional encoders use three layers with hidden dimension of the $\FF$ operation set to 2048, and dropout rate set to 0.1. LSTM-based encoders use 2 layers. We follow \newcite{Sperber2017} to tokenize and lowercase data, remove punctuation, and replace singletons with a special \texttt{unk} token. Beam size is set to 8.

For training, we find it important to pretrain on sequential data and finetune on lattice data (\cref{sec:exp-pretraining}).
This is in line with prior work \cite{Sperber2017} and likely owed to the fact that the lattices in this dataset are rather noisy, hampering training especially during the early stages.
We use Adam for training \cite{Kingma2014}. For sequential pretraining, we follow the learning schedule with warm-up and decay of \newcite{Vaswani2017}. Finetuning was sometimes unstable, so we finetune both using the warm-up/decay strategy and using a fixed learning rate of 0.0001 and report the better result. We use large-batch training with minibatch size of 1024 sentences, accumulated over 16 batched computations of 64 sentences each, due to memory constraints. Early stopping is applied when the BLEU score on a held-out validation set does not improve over 15 epochs, and the model with the highest validation BLEU score is kept.

\subsection{Main Results}
\label{sec:main_results}

\begin{table}[tb]
\centering
\begin{tabular}{@{}llll@{}}
\toprule
Encoder model                           & Inputs            & Fisher              & Callh.     \\ 
\midrule 
LSTM\footnote{BLEU scores taken from \newcite{Sperber2017}.\label{foot:latt}} & 1-best  & 35.9 & 11.8       \\ 
Seq.\ SA                                     & 1-best             & 35.71             &  12.36    \\ 
Seq.\ SA (directional)                 & 1-best            & 37.42              &  13.00    \\ 
\midrule 
Graph attention                           & lattice             & 35.71             & 11.87   \\  
LatticeLSTM\textsuperscript{\ref{foot:latt}} & lattice   & 38.0         &  14.1  \\ 
\midrule 
Lattice SA (proposed)                  & lattice              & \textbf{38.73} & \textbf{14.74}   \\  
\bottomrule
\end{tabular}
\caption{BLEU scores on Fisher (4 references) and Callhome (1 reference), for proposed method and several baselines. 
}
\label{tab:main_res}
\end{table}

Table~\ref{tab:main_res} compares our model against several baselines. Lattice models tested on Callhome are pretrained on Fisher and finetuned on Callhome lattices (Fisher+Callhome setting), while lattice models tested on Fisher use a Fisher+Fisher training setting. All sequential baselines are trained on the reference transcripts of Fisher. The first set of baselines operates on 1-best (sequential) inputs and includes a bidirectional LSTM, an unmasked self-attentional encoder (SA) of otherwise identical architecture with our proposed model, and a variant with directional masks \cite{Shen2018}. 
Next, we include a graph-attentional model that masks all but adjacent lattice nodes \cite{Velickovic2018} but is otherwise identical to the proposed model, and a LatticeLSTM. Note that these lattice models both use the cross-attention lattice-score bias (\cref{sec:enc-dec}).

Results show that our proposed model outperforms all examined baselines. Compared to the sequential self-attentional model, our models improves by 1.31--1.74 BLEU points. Compared to the LatticeLSTM, our model improves results by 0.64--0.73 BLEU points, while at the same time being more computationally efficient (\cref{sec:experiments-speed}). Graph attention is not able to improve over the sequential baselines on our task due to its restriction to local context.

\subsection{Computation Speed}
\label{sec:experiments-speed}

\begin{table}[tb]
\centering
\begin{tabular}{@{}lcccc@{}}
\toprule
                        & \multicolumn{2}{c}{Training} & \multicolumn{2}{c}{Inference}   \\
Encoder           & Batching &     Speed    & Batching  & Speed  \\ 
\midrule
\multicolumn{5}{c}{\textit{Sequential encoder models}} \\
LSTM              & \texttt{M}   &   4629        & --            &  715     \\ 
SA                   & \texttt{M}   & \textbf{5021}  & --            &  \textbf{796}      \\ 
\midrule
\multicolumn{5}{c}{\textit{LatticeLSTM and lattice SA encoders}} \\
LSTM              & --                &  178          & --            &  391     \\ 
LSTM              & \texttt{A}     &  710          & \texttt{A} &  538     \\ 
SA                   &  \texttt{M}   &\textbf{2963}& --         &  687     \\ 
SA                   &  \texttt{A}    &  748          & \texttt{A} &  \textbf{718}    \\ 
\bottomrule
\end{tabular}
\caption{Computation speed (words/sec), averaged over 3 runs. Batching is conducted manually (\texttt{M}), through autobatching (\texttt{A}), or disabled (--). The self-attentional lattice model displays superior speed despite using 3 encoder layers, compared to 2 layers for the LSTM-based models.}
\label{tab:speed}
\end{table}

The self-attentional lattice model was motivated not only by promising model accuracy (as confirmed above), but also by potential speed gains. We therefore test computation speed for training and inference, comparing against LSTM- and LatticeLSTM-based models. For fair comparison, we use a reimplementation of the LatticeLSTM so that all models are run with the exact same toolkits and identical decoder architectures. Again, LSTM-based models have two encoder layers, while self-attentional models have three layers. LatticeLSTMs are difficult to speed up through manually implemented batched computations, but similar models have been reported to strongly benefit from autobatching \cite{Neubig2017a} which automatically finds operations that can be grouped after the computation graph has been defined. Autobatching is implemented in DyNet but not available in many other deep learning toolkits, so we test both with and without autobatching. Training computations are manually or automatically batched across 64 parallel sentences, while inference speed is tested for single sentences with forced decoding of gold translations and without beam search. We test with DyNet commit \texttt{8260090} on an Nvidia Titan Xp GPU and average results over three runs.

 Table~\ref{tab:speed} shows the results. For sequential inputs, the self-attentional model is slightly faster than the LSTM-based model. The difference is perhaps smaller than expected, which can be explained by the larger number of layers in the self-attentional model, and the relatively short sentences of the Fisher corpus that reduce the positive effect of parallel computation across sequence positions. For lattice-based inputs, we can see a large speed-up of the self-attentional approach when no autobatching is used. Replacing manual batching with autobatching during training for the self-attentional model yields no benefits. Enabling autobatching at inference time provides some speed-up for both models. Overall, the speed advantage of the self-attentional approach is still very visible even with autobatching available.

\subsection{Feature Ablation}
\label{sec:ablation}

\begin{table}[tb]
\centering
\begin{tabular}{@{}ccccrr@{}}
\toprule
\begin{tabular}[c]{@{}c@{}}reachability\\ mask\end{tabular} & dir. & prob. & \begin{tabular}[c]{@{}l@{}}latt.\\ pos.\end{tabular} & Fisher & Callh.  \\ \midrule
\checkmark          & \checkmark  & \checkmark   & \checkmark            & \textbf{38.73}  &  \textbf{14.74} \\ 
\midrule
\checkmark           & \checkmark  & \checkmark   &                               & 38.25      & 12.45 \\  
\checkmark          & \checkmark  &                       & \checkmark            & 37.52      &  14.37 \\ 
\checkmark          &                      & \checkmark   & \checkmark            & 35.49      & 12.83 \\ 
\midrule
                             & \checkmark   &                       &                               & 30.58      & 9.41 \\  
 \bottomrule
\end{tabular}
\caption{Ablation over proposed features, including reachability masks, directional (vs.\ non-directional) masking, probabilistic (vs.\ binary) masking, and lattice positions (vs.\ topological positions).}
\label{tab:ablation}
\end{table}

We next conduct a feature ablation to examine the individual effect of the improvements introduced in \cref{sec:proposed_model}. Table~\ref{tab:ablation} shows that longest-path position encoding outperforms topological positions, the probabilistic approach outperforms binary reachability masks, and modeling forward and reversed lattices with separate attention heads outperforms the non-directional approach. Consistently with the findings by \newcite{Sperber2017}, lattice scores are more effectively exploited on Fisher than on Callhome as a result of the poor lattice quality for the latter. The experiment in the last row demonstrates the effect of keeping the lattice contents but removing all structural information, by rearranging nodes in linear, arbitrary topological order, and applying the best sequential model. Results are poor and structural information clearly beneficial.

\subsection{Behavior At Test Time}

\begin{table}[tb]
\centering
\begin{tabular}{@{}lccc@{}}
\toprule
                   &  Lattice oracle & 1-best    & Lattice \\ \midrule
\multicolumn{4}{c}{\textit{Fisher}} \\
Sequential SA &          47.84         & 37.42       &  --              \\ 
Lattice SA     & 47.69       & 37.56       & 38.73        \\ 
\midrule
\multicolumn{4}{c}{\textit{Callhome}} \\
Sequential SA          &  17.94      &  13.00      &  --                 \\ 
Lattice SA        & 18.54        & 13.90       & 14.74        \\ 
\bottomrule
\end{tabular}
\caption{Fisher and Callhome models, tested by inputting lattice oracle paths, 1-best paths, and full lattices.}
\label{tab:test-time}
\end{table}

To obtain a better understanding of the proposed model, we compare accuracies to the sequential self-attentional model when translating either lattice oracle paths, 1-best transcripts, or lattices. The lattice model translates sequences by treating them as lattices with only a single complete path and all transition probabilities set to 1. Table~\ref{tab:test-time} shows the results for the Fisher+Fisher model evaluated on Fisher test data, and for the Fisher+Callhome model evaluated on Callhome test data. We can see that the lattice model outperforms the sequential model even when translating sequential 1-best transcripts, indicating benefits perhaps due to more robustness or increased training data size for the lattice model. However, the largest gains stem from using lattices at test time, indicating that our model is able to exploit the actual test-time lattices. Note that there is still a considerable gap to the translation of lattice oracles which form a top-line to our experiments.

\subsection{Effect of Pretraining and Finetuning}
\label{sec:exp-pretraining}

Finally, we analyze the importance of our strategy of pretraining on clean sequential data before finetuning on lattice data. Table~\ref{tab:callhome} shows the results for several combinations of pretraining and finetuning data.
The first thing to notice is that pretraining is critical for good results. Skipping pretraining performs extremely poorly, while pretraining on the much smaller Callhome data yields results no better than the sequential baselines (\cref{sec:main_results}). We conjecture that pretraining is beneficial mainly due to the rather noisy lattice training data, while for tasks with cleaner training lattices pretraining may play a less critical role.

The second observation is that for the finetuning stage, domain appears more important than data size: Finetuning on Fisher works best when testing on Fisher, while finetuning on Callhome works best when testing on Callhome, despite the Callhome finetuning data being an order of magnitude smaller.
 This is encouraging, because the collection of large amounts of training lattices can be difficult in practice.

\begin{table}[tb]
\centering
\begin{tabular}{@{}cccc@{}}
\toprule
Sequential data  & Lattice data          & Fisher               & Callh.  \\ 
\midrule
--                         & Fisher                  & 1.45                  & 1.78              \\ 
Callhome	           & Fisher		         & 34.52                & 13.04              \\ 
Fisher	          & Callhome             &  35.47                & \textbf{14.74}   \\ 
Fisher	          & Fisher		        &  \textbf{38.73}    & 14.59               \\ 
\bottomrule
\end{tabular}
\caption{BLEU scores for several combinations of Fisher (138k sentences) and Callhome (15k sentences) training data.}
\label{tab:callhome}
\end{table}

\section{Related Work}
The translation of lattices rather than sequences has been investigated with traditional machine translation models \cite{Ney1999,Casacuberta2004,Saleem2004,Zhang2005,Matusov2008,Dyer2008}, but these approaches rely on independence assumptions in the decoding process that no longer hold for neural encoder-decoder models. Neural lattice-to-sequence models were proposed by \newcite{Su2017,Sperber2017}, with promising results but slow computation speeds. Other related work includes gated graph neural networks \cite{Li2016a,Beck2018}. As an alternative to these RNN-based models, GCNs have been investigated \cite{Duvenaud2015,Defferrard2016,kearnes2016molecular,kipf2016semi}, and used for devising tree-to-sequence models \cite{Bastings2017,Marcheggiani2018}. We are not aware of any application of GCNs to lattice modeling. Unlike our approach, GCNs consider only local context, must be combined with slower LSTM layers for good performance, and lack support for lattice scores.

Our model builds on previous works on self-attentional models \cite{cheng2016a,Parikh2016,Lin2017,Vaswani2017}. The idea of masking has been used for various purposes, including occlusion of future information during training \cite{Vaswani2017}, introducing directionality \cite{Shen2018} with good results for machine translation confirmed by \newcite{Song2018}, and soft masking \cite{Im2017,Sperber2018}. The only extension of self-attention beyond sequence modeling we are aware of is graph attention \cite{Velickovic2018} which uses only local context and is outperformed by our model.

\section{Conclusion}
This work extended existing sequential self-attentional models to lattice inputs, which have been useful for various purposes in the past. We achieve this by introducing probabilistic reachability masks and lattice positional encodings. Experiments in a speech translation task show that our method outperforms previous approaches and is much faster than RNN-based alternatives in both training and inference settings. Promising future work includes extension  to tree-structured inputs and application to other tasks.


\section*{Acknowledgments}
The work leading to these results has received funding from the European Union under grant agreement no 825460.

\bibliography{library}
\bibliographystyle{acl_natbib}


\clearpage
\appendix
\section{Path Duplication Invariance}
\label{app:duplication-invariance}

\begin{figure}[tb]
\centering
\includegraphics[width=8cm]{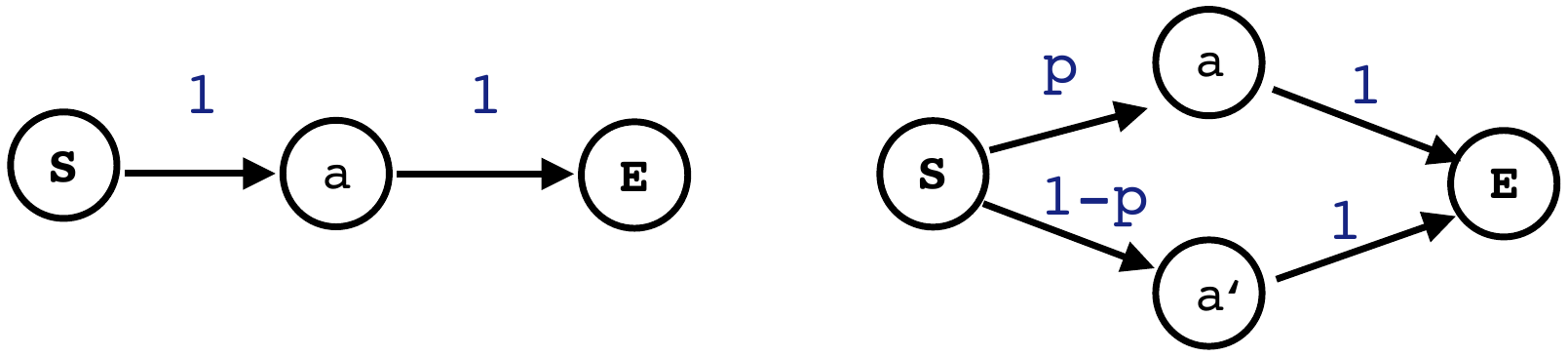}
{\normalsize
\begin{tabular}{c|ccccccccc}
sequential  & \texttt{S} & \texttt{a}   & \texttt{E} \\
\midrule
\texttt{S}         & 1            & 1            & 1       \\
\texttt{a}         & 0            & 1               & 1          \\
\texttt{E}         & 0            & 0               & 1
\end{tabular}
\begin{tabular}{c|ccccccccc}
duplicated     & \texttt{S} & \texttt{a}    & \texttt{a'}   & \texttt{E} \\
\midrule
\texttt{S}        & 1            & $p$           &  $(1-p)$     & 1 \\
\texttt{a}        & 0            & 1               &  0               & 1 \\
\texttt{a'}        & 0            & 0               &  1              & 1 \\
\texttt{E}        & 0            & 0               &   0              & 1
\end{tabular}
}
\caption{A sequential lattice, and a variant with a duplicated path, where nodes \texttt{a} and \texttt{a'} are labeled with the same word token. The matrices contain pairwise reaching probabilities in forward direction, where rows are queries, columns are keys.}
\label{fig:duplication-invariance}
\end{figure}

Figure~\ref{fig:duplication-invariance} shows a sequential lattice, and a lattice derived from it but with a duplicated path. Semantically, both are equivalent, and should therefore result in identical neural representations. Note that while in practice duplicated paths should not occur, paths with partial overlap are quite frequent. It is therefore instructive to consider this hypothetical situation. Below, we demonstrate that the binary masking approach (\cref{sec:bin-masks}) is biased such that computed representations are impacted by path duplication. In contrast, the probabilistic approach (\cref{sec:prob_masks}) is invariant to path duplication.

We consider the example of Figure~\ref{fig:duplication-invariance}, discussing only the forward direction, because the lattice is symmetric and computations for the backward direction are identical. We follow notation of Equations~\ref{eq:background-sa-score} through \ref{eq:weighted_sum}, using $\langle \text{\texttt{a}},\text{\texttt{b}}\rangle$ as abbrevation for $f\left(q\left(\mathbf{x}_\text{\texttt{a}}\right),k\left(\mathbf{x}_\text{\texttt{b}}\right)\right)$ and $\mathbf{v}_\text{\texttt{a}}$ to abbreviate $v(\mathbf{x}_\text{\texttt{a}})$. Let us consider the computed representation for the node \texttt{S} as query. For the sequential lattice with binary mask, it is:
\begin{align}
\mathbf{y}_\text{\texttt{S}} =    \frac{1}{C}\left(    e^{\langle\text{\texttt{S}},\text{\texttt{S}}\rangle} \mathbf{v}_\text{\texttt{S}}  +    e^{\langle\text{\texttt{S}},\text{\texttt{a}}\rangle} \mathbf{v}_\text{\texttt{a}}  +   e^{\langle\text{\texttt{S}},\text{\texttt{b}}\rangle} \mathbf{v}_\text{\texttt{b}}      \right) \label{eq:appendix_seq}
\end{align}
Here, $C$ is the softmax normalization term that ensures that exponentiated similarities sum up to 1.

In contrast, the lattice with duplication results in a doubled influence of $\mathbf{v}_\text{\texttt{a}}$:

\begin{align}
\mathbf{y}_\text{\texttt{S}} =    &   \frac{1}{C}\Big(    e^{\langle\text{\texttt{S}},\text{\texttt{S}}\rangle} \mathbf{v}_\text{\texttt{S}}  +    e^{\langle\text{\texttt{S}},\text{\texttt{a}}\rangle} \mathbf{v}_\text{\texttt{a}}  \nonumber \\
    &  +    e^{\langle\text{\texttt{S}},\text{\texttt{a'}}\rangle} \mathbf{v}_\text{\texttt{a'}}    +      e^{\langle\text{\texttt{S}},\text{\texttt{E}}\rangle} \mathbf{v}_\text{\texttt{E}}      \Big)\nonumber \\  
                                           =    &   \frac{1}{C}\left( e^{\langle\text{\texttt{S}},\text{\texttt{S}}\rangle} \mathbf{v}_\text{\texttt{S}}  +    2e^{\langle\text{\texttt{S}},\text{\texttt{a}}\rangle} \mathbf{v}_\text{\texttt{a}}   + e^{\langle\text{\texttt{S}},\text{\texttt{E}}\rangle} \mathbf{v}_\text{\texttt{E}}      \right).\nonumber 
\end{align}

The probabilistic approach yields the same result as the binary approach for the sequential lattice (Equation~\ref{eq:appendix_seq}). For the lattice with path duplication, the representation for the node \texttt{S} is computed as follows:
\begin{align}
\mathbf{y}_\text{\texttt{S}} =    &   \frac{1}{C}\Big(    e^{\langle\text{\texttt{S}},\text{\texttt{S}}\rangle} \mathbf{v}_\text{\texttt{S}}  +    e^{\langle\text{\texttt{S}},\text{\texttt{a}}\rangle+\log p} \mathbf{v}_\text{\texttt{a}}  \nonumber \\
    &  +    e^{\langle\text{\texttt{S}},\text{\texttt{a'}}\rangle+\log(1-p)} \mathbf{v}_\text{\texttt{a'}}    +      e^{\langle\text{\texttt{S}},\text{\texttt{E}}\rangle} \mathbf{v}_\text{\texttt{E}}      \Big)\nonumber \\  
                                           =    &   \frac{1}{C}\Big(    e^{\langle\text{\texttt{S}},\text{\texttt{S}}\rangle} \mathbf{v}_\text{\texttt{S}}  +    e^{\langle\text{\texttt{S}},\text{\texttt{a}}\rangle}e^{\log p} \mathbf{v}_\text{\texttt{a}}  \nonumber \\
    &  +    e^{\langle\text{\texttt{S}},\text{\texttt{a'}}\rangle}e^{\log(1-p)} \mathbf{v}_\text{\texttt{a'}}    +      e^{\langle\text{\texttt{S}},\text{\texttt{E}}\rangle} \mathbf{v}_\text{\texttt{E}}      \Big)\nonumber \\  
                                           =    &   \frac{1}{C}\Big(    e^{\langle\text{\texttt{S}},\text{\texttt{S}}\rangle} \mathbf{v}_\text{\texttt{S}}  +    pe^{\langle\text{\texttt{S}},\text{\texttt{a}}\rangle} \mathbf{v}_\text{\texttt{a}}  \nonumber \\
    &  +    (1-p)e^{\langle\text{\texttt{S}},\text{\texttt{a'}}\rangle} \mathbf{v}_\text{\texttt{a'}}    +      e^{\langle\text{\texttt{S}},\text{\texttt{E}}\rangle} \mathbf{v}_\text{\texttt{E}}      \Big)\nonumber \\  
                                           =    &   \frac{1}{C}\Big(    e^{\langle\text{\texttt{S}},\text{\texttt{S}}\rangle} \mathbf{v}_\text{\texttt{S}}  +    e^{\langle\text{\texttt{S}},\text{\texttt{a}}\rangle} \mathbf{v}_\text{\texttt{a}}    +      e^{\langle\text{\texttt{S}},\text{\texttt{E}}\rangle} \mathbf{v}_\text{\texttt{E}}      \Big).\nonumber 
\end{align}

The result is the same as in the semantically equivalent sequential case (Equation~\ref{eq:appendix_seq}), the computation is therefore invariant to path duplication. The same argument can be extended to other queries, to other lattices with duplicated paths, as well as to the lattice-biased encoder-decoder attention.

\section{Qualitative Analysis}
\label{app:qualitative}

We conduct a manual inspection and showcase several common patterns in which the lattice input helps improve translation quality, as well as one counter example. In particular, we compare the outputs of the sequential and lattice models according to the 3rd and the last row in Table~\ref{tab:main_res}, on Fisher.

\subsection{Example 1}
In this example, the ASR 1-best contains a bad word choice ({\it quedar} instead of {\it qué tal}). The correct word is in the lattice, and can be disambiguated by exploiting long-range self-attentional encoder context.
\begin{description}
  \item[gold transcript:] {\it Qué tal, eh, yo soy Guillermo, ¿Cómo estás?}
  \item[ASR 1-best:] {\it quedar eh yo soy guillermo cómo estás}
  \item[seq2seq output:] {\it \underline{stay} eh i ' m guillermo how are you}
  \item[ASR lattice:] \begin{minipage}{\linewidth}
\includegraphics[width=5.0cm]{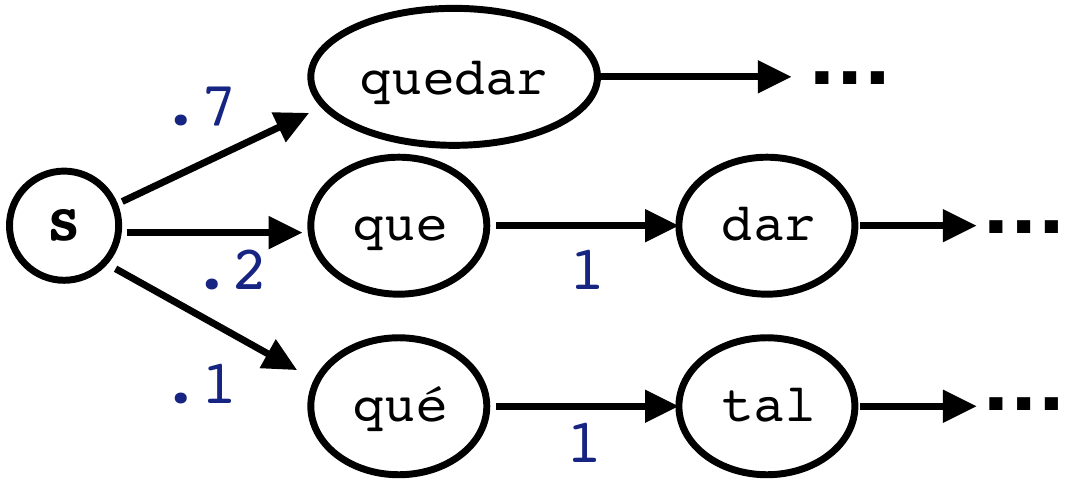}
 \end{minipage}
  \item[lat2seq output:] {\it \underline{how are you} eh i ' m guillermo how are you}
\end{description}

\subsection{Example 2}
Here, the correct word {\it graduar} does not appear in the lattice, instead the lattice offers many incorrect alternatives of high uncertainty. The translation model evidently goes with a linguistically plausible guess, ignoring the source side.
\begin{description}
  \item[gold transcript:] {\it Claro Es, eh, eh, o sea, yo me, me voy a graduar con un título de esta universidad.}
  \item[ASR 1-best:] {\it claro existe eh o sea yo me me puedo habar con un título esta universidad}
  \item[seq2seq output:] {\it sure it exists i mean i can \underline{talk} with a title}
  \item[ASR lattice:] \begin{minipage}{\linewidth}
\includegraphics[width=5.0cm]{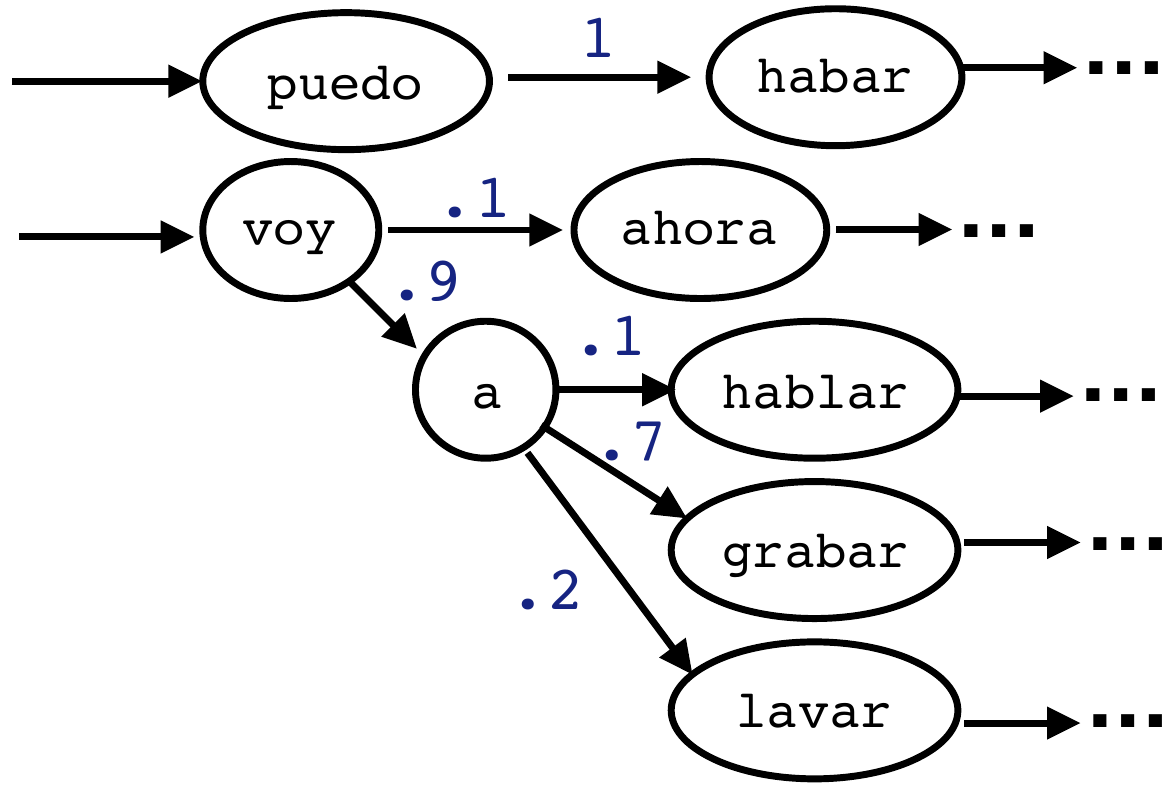}
 \end{minipage}
  \item[lat2seq output:] {\it sure i mean i ' m going to \underline{take} a university title}
\end{description}

\subsection{Example 3}

In this example, {\it o sea} ({\it I mean}) appears with slightly lower confidence than {\it saben} ({\it they know}), but is chosen for a more natural sounding target sentence

\begin{description}
  \item[gold transcript:] {\it No, o sea, eso es eh, clarísimo para mi}
  \item[ASR 1-best:] {\it no saben eso es eh clarísimo para mi}
  \item[seq2seq output:] {\it \underline{they don ' t know} that ' s eh sure for me}
  \item[ASR lattice:] \begin{minipage}{\linewidth}
\includegraphics[width=5.5cm]{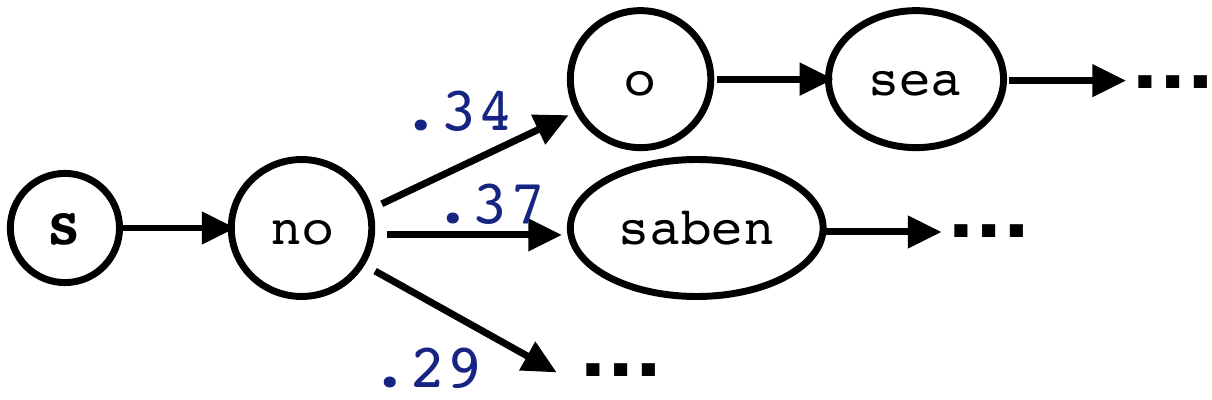}
 \end{minipage}
  \item[lat2seq output:] {\it no \underline{i mean} that ' s very clear for me}
\end{description}

\subsection{Counter Example}

In this counter example, the translation model gets confused from the additional and wrong lattice context and no longer produces the correct output.

\begin{description}
  \item[gold transcript:] {\it sí}
  \item[ASR 1-best:] {\it sí}
  \item[seq2seq output:] {\it yes}
  \item[ASR lattice:] \begin{minipage}{\linewidth}
\includegraphics[width=3.5cm]{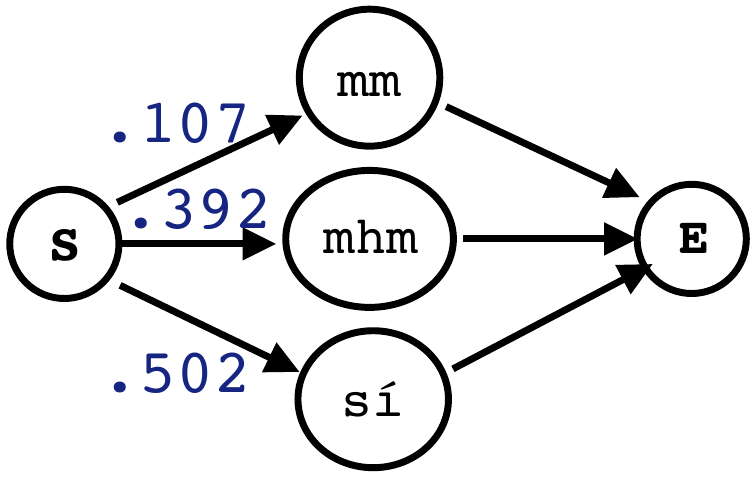}
 \end{minipage}
  \item[lat2seq output:] {\it mm}
\end{description}

\end{document}